\pdfoutput=1
\documentclass[11pt]{article}

\usepackage{acl}
\usepackage{times}
\usepackage{latexsym}

\usepackage[T1]{fontenc}

\usepackage[utf8]{inputenc}

\usepackage{microtype}

\usepackage{hyperref}       % hyperlinks
\usepackage{url}            % simple URL typesetting
\usepackage{booktabs}       % professional-quality tables see https://www.ctan.org/pkg/booktabs
\usepackage{amsfonts}       % blackboard math symbols
\usepackage{nicefrac}       % compact symbols for 1/2, etc.
\usepackage{graphicx}

\usepackage{natbib}
\usepackage{doi}

\title{Misogyny classification of German newspaper forum comments} 
\author{Johann Petrak \and Brigitte Krenn\\
  Austrian Research Instititute \\
  for Artificial Intelligence \\
  Vienna, Austria \\
  }
  
\hypersetup{
pdftitle={Misogyny classification of German newspaper forum comments},
}

\begin{document}
\maketitle

\begin{abstract}
	This paper presents work 
	on detecting miso\-gyny in the comments 
	of a large Austrian German language newspaper forum. We describe the creation of a 
    corpus of 6600 comments which were annotated with 5 levels of misogyny. The forum
    moderators were involved as experts in the creation of the annotation guidelines and the 
    annotation of the comments. 
	We also describe the results of training transformer-based classification models 
	for both binarized and original label classification of that corpus. 
\end{abstract}

\section{Introduction and Motivation}
\label{sec:introduction}

The ever more widespread use of social media and user-contributed content 
also causes  an increase of toxic or offensive language and other forms of unwanted contributions which may need to get detected and removed. In this paper, we present work aimed at supporting  moderators
of a large daily Austrian (German language) newspaper which allows registered users to discuss the articles published on its  web-site. Users post some 20K to 50K comments per day. An analysis of commenting behaviour has shown that only a third of the users participating in the online discussion are women and that one important reason why women avoid participating in article forum discussions is the presence of sexist comments. The aim therefore was to use automatic classification of sexist/misogynist comments to support moderators in detecting such comments  in order to  provide a more welcoming and safer climate of discussion especially for female users. 

For this, a corpus of 6600 comments was collected and annotated and subsequently used to train a classifier used to flag comments or entire discussion forums with a high number of suspected misogynist comments.

\section{Related Work}
\label{sec:relatedwork}

Together with work on toxic and offensive language classification in recent years, there has 
also been increasing work on the classification of sexist or misogynist language, sometimes as part of 
a more general toxic language classification task. \citet{Hewitt2016} give an overview over earlier work and describe a dataset of Tweets containing abusive sexist terms. 
\citet{Anzovino2018} present work on creating a dataset of tweets, subsequently used as part of IberEval-2018 and EvalIta-2018 challenges \cite{Fersini2018} for sexism classification.
\citet{Shushkevich2019} give an overview over misogyny detection in social media, specifically Twitter. 

\citet{Waseem2016} describe work on a corpus of 16K tweets for detecting toxic and hate speech including sexist slurs or defending sexism (3383 tweets with sexist content). 
\citet{Frenda2019} present work on the datasets described in \cite{Fersini2018} and \cite{Waseem2016}. \citet{Sharifirad2019} include some more detailed description of sexist language and is based on another dataset of English language tweets. \citet{Parikh2019} describe work on categorizing accounts of sexism from the Everyday Sexism Project website through fine-grained multilabel classification. Other datasets are described or used in \cite{Chiril2020a}  (12K tweets in French) and \cite{Grosz2020} (tweets, work-related quotes, press quotes and other sources), \cite{Bhattacharya2020a} and \cite{SafiSamghabadi2020} (Youtube comments in Indian English, Hindi and Bengla), \cite{RodriguezSanchez2020} (Spanish language tweets) and \cite{Zeinert2021}  (Danish language dataset sampled from several social media sites). 

The EXIST task at IberLEF 2021 \cite{RodriguezSachez2021} addresses the identification and categorization of sexism in English and Spanish language tweets and postings from \url{Gab.com}. Most recently, SemEval 2023 Task 10\footnote{\url{https://codalab.lisn.upsaclay.fr/competitions/7124}} provides an English language corpus of 20000 texts sampled from Gab and Reddit, annotated with 3 hierarchical labels. 
%BK: im obgen Text wird of present verwendet. Sollten wir umschreiben, wenn genug Zeit da ist.

% The social media datasets described in those works are all collected primarily by searching for  a set of keywords related to sexism or misogyny. In some cases the initial gathering of texts is then  extended by identifying and gathering additional data from accounts identified to post sexist content. This form of data collection introduces at least some bias towards a specific form of sexism related to those initial keywords. The corpus presented in this paper is an attempt to collect a wider range of texts which also include subtle or implicit forms of sexism which may also be challenging to judge by humans. It is, to the best of our knowledge, also the first corpus on sexism/misogyny created for German. 

\section{Corpus Creation}
\label{sec:corpuscreation}

The aim of manual corpus annotation was to reflect the judgement of moderators in their 
everyday work. For this reason, the manual annotations were carried out by 8 annotators of 
which 7 were experienced moderators. The 8th annotator is a natural language processing and corpus linguistics expert. One of the annotators is among the authors of this paper. There were 3 male and 5 female annotators. 

%BK Frage: finden sich Unterschiede in den Bewertungen von m/f Annotators?

Since the phenomenon of "sexism"/"misogyny" is complex, guidelines were created to 
describe the most important kinds of sexism relevant for the annotation task. For this
we used the categorization from \cite{Parikh2019} as inspiration. The guidelines also 
attempt to clarify some of the difficulties likely to be encountered: how to decide if 
there is not enough context, what if the sexist remark is aimed at a man or men in general, 
how to treat "reported sexism" \cite{Chiril2020}. However, the guidelines follow the 
aim of providing help for annotating in a way that reflects the daily work of moderators and the newspaper's editorial concept. They are not meant as an accurate abstract definition of sexism and misogyny. The (German language) guidelines are available online\footnote{\url{https://ofai.github.io/femdwell/annotation_guidelines_v1.pdf}}. We will refer to the task as "misogyny classification" in the rest of the paper. 

%BK: todo: add references from the theoretical literature on sexism/misogyny.

Postings were annotated by assigning one of 5 possible labels 0 .. 4, corresponding to 
0=absence of misogyny and 4 levels of "severity" of the expressed misogyny as perceived by the individual annotators, with 1=mild, 2=present, 3=strong, 4=extreme. 
This was done on the one hand to reflect the personal aspect in the assessment of misogyny, and on the other hand to identify the instances with the biggest disagreements among annotators.

Postings to be annotated were collected from several different sources: (1) a collection of 
postings which had been reported with a (free text) reporting reason that included a keyword
related to sexism/misogyny, (2) postings which were reported %but 
with a different reason, (3) postings randomly sampled from all available postings, (4) a subset of postings (2) preclassified with an early version of the binary classifier trained on the first 2800 annotated postings and (5) postings from 24 article forums which have been identified to contain an above-average number of postings considered sexist, preclassified with the same  early binary classifier. Preclassified postings were selected from the highest probability label "1" postings (to correct false positives) as well as those label "0" and "1" postings with close to 0.5 probability  (to add what may be hard to classify instances). 

Postings were given to annotators in batches of 100, by creating a spreadsheet from the posting texts and preparing a selection field for selecting one of the 5 possible labels. The first batch of 100 postings was given to all 8 annotators and then analysed to find postings with the biggest
disagreement: for this we calculated a heuristic disagreement score based on all 
pairwise distances between the labels, where the distance between labels 0 and 1 were 
defined to be 4, and distances between labels $l_i, l_j >0$ defined to be $|l_i-l_j|$. Examples with high scores were then discussed among annotators to clarify the annotation guidelines or clear up misunderstandings. 

After this, each round of 100 postings was given to a random selection of 3 annotators available at the time. This was done in order to compromise between annotating as many postings as possible given the available time and resources and still get enough annotators for each posting to identify disagreements. In total, 66 rounds with 6600 postings were annotated (20300 annotations). The overall distribution of assigned labels is shown in Table~\ref{table:dist5}. 

\begin{table}[htbp!]
	\caption{Distribution of assigned labels}
	\centering
	\begin{tabular}{lrrrrrr}
		\toprule
		Label & 0 & 1 & 2 & 3 & 4 & 1...4     \\
		\midrule
		\%    & 66.5 & 7.3 & 14.2 & 9.4 & 2.6 & 33.5 \\
		\bottomrule
	\end{tabular}
	\label{table:dist5}
\end{table}

% Some text on the definition of sexism/misogyny used for this work. Some words on the annotation guidelines.
%Mention that there many corpora available on sexism classification do not include detailed annotation guidelines. Mention that our guidelines are partly based on the categories listed in \cite{Parikh2019} and \cite{Anzovino2018}

\section{Annotator Agreement and Corpus Analysis}
\label{sec:corpusanalysis}

Krippendorff Alpha over all annotations was 0.36 (nominal scale) and 0.64 (ordinal scale). After binarization of the 5 possible annotations into 0 (for no sexism) and 1 (for labels $1\dots 4$), Krippendorff Alpha was 0.60. 
%BK: Bei Krippendorff Alpha tue ich mir schwer: wir haben 0.64 overall (ordinal) und binary nur 0.60. Wie ist das zu interpretieren?
Overall agreement was 0.65 / 0.83 (binary) if macro averaged over all agreements of pairs of users, and 0.66 / 0.82 (binary) if micro averaged over all pairs of annotations. F1.0 macro over all 22300 pairs
of annotations was 0.390 / 0.803 (binary). Overall Cohen's Kappa was 0.39 / 0.63 (binary) if macro averaged over all kappas for pairs of users. This indicates that there was considerable difference of opinion or difficulty in assigning the fine-grained labels. 

As shown in Table~\ref{table:dist5} the overall rate of assigned positive labels (classes $1\dots 4$) was 0.335. Looking at all pairs of annotations in the dataset the relative frequencies of annotation pairs (confusion matrix) is shown in Table~\ref{table:conf1}.  This illustrates the large rate of disagreement among annotators, especially on estimating the fine-grained degree of mysogyny (labels $1\dots 4$). 
% better explanation of confusion matrix

\begin{table}[htbp!]
	\caption{Relative frequence of annotation pairings}
	\centering
	\begin{tabular}{l|lllll}
		\toprule
		  & 0 & 1 & 2 & 3 & 4 \\
		\midrule
		0 & 0.525 & 0.032 & 0.037 & 0.015 & 0.003 \\
		1 & 0.032 & 0.014 & 0.020 & 0.009 & 0.001 \\
		2 & 0.037 & 0.020 & 0.052 & 0.036 & 0.007 \\
		3 & 0.015 & 0.009 & 0.036 & 0.044 & 0.016 \\
		4 & 0.003 & 0.001 & 0.007 & 0.016 & 0.013 \\
		\bottomrule
	\end{tabular}
	\label{table:conf1}
\end{table}

%BK: tue mir schwer bei der Interpretation der Confusion-Matrix

%%% TODO: example texts? Used to clear up anno guidelines, example hard texts
%%% TODO: distribution, stats for our own disagreement score? Per round/source analyses of 
%% IAA, disagreement scores. 
%%% Reiterate: first German corpus, first corpus which tries to deliberately include "hard" cases 
%%% of subtle sexism etc. 

We deliberately did not decide on a single "best" way to decide on a final single label as
the judgement on misogyny may depend on personal opinion and we believe it would be wrong to 
assume there is a single correct value for each instance. Other instances may have received 
labels by mistake (e.g. misinterpretation of the text or of the annotation guidelines). For creating 
a training set we used different strategies to resolve disagreements (see Section~\ref{sec:model}).

We aim to make the corpus available for academic research, the necessary steps for doing this legally and in accordance with Austrian and EU data protection law are still ongoing.  
% with the hope that it will be used not only for replicating and extending the work on creating useful classification models from it, but also to update or add to the labels or extend the corpus with additional texts in order to better understand how people may disagree on what actually constitutes sexism and misogyny. For this reason it is made available as a git repository which is meant to receive 
%updates and fixes by contributors and see new releases in the future. The corpus contains the text of the postings (with no information to identify the author) as well as all original labels (with no information about the annotator except their gender) and the identification of the source of the posting.
%BK: was ist mit source of the posting gemeint?

This is not only the first German language corpus related to misogyny detection but also the first corpus in any language which covers such a wide range of often very subtle ways a comment may express mysogyny, and which contains many comments where human annotators will have different opinions on which label is most adequate. 
% We hope it will therefore also be useful for further research into how to deal with ambiguity and  the lack of a single ground truth in text classification. 

\section{Sexism Classification Model}
\label{sec:model}

The main purpose of the deployed classification model is to alert moderators both of individual
sexist comments and article forums with a high rate of potentially sexist comments. For this 
reason, our main interest is a binary classifier (original label 0 vs original labels 1 to 4). 
However we also studied the performance of a model for predicting the original label, both 
as seen as a multiclass classification task and as an ordinal regression task.  Finally, we
investigated if combining both the binary and multiclass tasks into a multi-task model would
impair or improve the performance of the individual tasks. 

All models are based on a transformer architecture \cite{Vaswani2017a} with one or two classification
heads on top of the pooling layer. We used the pretrained German BERT models \texttt{gbert-base}\footnote{\url{https://huggingface.co/deepset/gbert-base/tree/main}} and 
\texttt{gbert-large}\footnote{\url{https://huggingface.co/deepset/gbert-large/tree/main}} \cite{Chan2020}.

For all models, accuracy and F1.0-macro metrics where estimated using 5-fold cross-validation 
with class-stratified folds. 10\% of the training set were used as dev-set and this split 
was class-stratified as well. Hyper-parameters where selected in a step-wise process where manually chosen values for a few hyper-parameters were evaluated using grid search. For many parameters, however, there was no clear best selection as the range of F1.0-macro estimation results  introduced by different random seeds was larger or comparable to the changes in F1.0-macro estimation for different parameter values. In such cases we chose an intermediate value among those with similar estimates. The final set of hyper-parameters used for all models described below was:
BERT maximum sequence length 192, batch size 8, gradient accumulation: 1 batch, learning rate
7.5e-06, language model dropout rate 0.1 and no layer-wise learning rate adjustment. The AdamW optimizer with weight decay 0.01 and linear warm-up during 200 training steps was used. For all classification 
heads we used an addition layer with 768 hidden units, ReLU and no dropout before the actual 
output layer. 

We evaluated the following single-head models: Bin (binary classification), Multi (multiclass classification), Coral (multiclass classification using an implementation of the CORAL ordinal regression model \cite{Cao2020}); and the following dual-head models: BinMulti (binary and multiclass
heads combined), BinCoral (binary and CORAL heads combined). For each of these 5 models we 
evaluated a variant based on the \texttt{bert-base} model (/B) and one based on the \texttt{bert-large} 
model (/L). 

Table~\ref{table:mostfreq} shows the estimation results (accuracy and F1.0-macro) on a training
set where the original and binarized targets where selected as the most frequently assigned labels,
falling back to the highest most frequently assigned label or the maximum assigned label. 
For dual-head multitask models there are two lines, showing the binary model as head 1 (:1) and 
the multiclass model as head 2 (:2). 

\begin{table}[htbp!]
\caption{Accuracy and F1.0 macro estimates ($\pm$ standard deviation) for models trained on the 
most frequent original (coral, multi) and binarized (bin) sexism label}
\label{table:mostfreq}
\centering
\begin{tabular}{lrr}
\toprule
Model& Accuracy & F1.0 macro \\
\midrule
Bin/B & 0.763$\pm$0.012 & 0.735$\pm$0.007 \\
Bin/L & 0.757$\pm$0.050 & 0.685$\pm$0.159 \\
Multi/B & 0.608$\pm$0.026 & \textbf{0.305}$\pm$0.013 \\
Multi/L & 0.640$\pm$0.043 & 0.262$\pm$0.093 \\
Coral/B & 0.651$\pm$0.019 & 0.281$\pm$0.011 \\
Coral/L & 0.662$\pm$0.017 & 0.250$\pm$0.082 \\
BinMulti/B:1 & 0.757$\pm$0.017 & 0.729$\pm$0.018 \\
BinMulti/B:2 & 0.612$\pm$0.015 & 0.293$\pm$0.016 \\
BinMulti/L:1 & 0.733$\pm$0.059 & 0.666$\pm$0.153 \\
BinMulti/L:2 & 0.506$\pm$0.213 & 0.255$\pm$0.121 \\
BinCoral/B:1 & 0.756$\pm$0.012 & 0.729$\pm$0.008 \\
BinCoral/B:2 & 0.656$\pm$0.013 & 0.269$\pm$0.016 \\
BinCoral/L:1 & \textbf{0.788}$\pm$0.015 & \textbf{0.763}$\pm$0.017 \\
BinCoral/L:2 & \textbf{0.658}$\pm$0.009 & 0.281$\pm$0.013 \\
\bottomrule
\end{tabular}
\end{table}

The dual-head model combining binary and CORAL ordinal regression heads shows the best 
accuracy estimates both for the binary and the multiclass tasks, and also the best 
F1.0-macro for the binary tasks while the performance of the base single-head multiclass 
model shows best F1.0-macro for the malticlass task. Interestingly, the models based on 
\texttt{bert-large} did not always perform better than those based on \texttt{bert-base}. 

\begin{table}[htbp!]
\caption{Accuracy and F1.0 macro estimates ($\pm$ standard deviation) for models trained on the 
	highest original (coral, multi) and binarized (bin) sexism label}
\label{table:choosemax}
\centering
\begin{tabular}{lrr}
\toprule
Model& Accuracy & F1.0 macro \\
\midrule
Bin/B & 0.760$\pm$0.008 & 0.760$\pm$0.008 \\
Bin/L & \textbf{0.777}$\pm$0.006 & \textbf{0.776}$\pm$0.005 \\
Multi/B & 0.509$\pm$0.010 & \textbf{0.353}$\pm$0.009 \\
Multi/L & 0.492$\pm$0.044 & 0.305$\pm$0.108 \\
Coral/B & 0.529$\pm$0.012 & 0.302$\pm$0.014 \\
Coral/L & 0.541$\pm$0.020 & 0.333$\pm$0.023 \\
BinMulti/B:1 & 0.762$\pm$0.009 & 0.762$\pm$0.009 \\
BinMulti/B:2 & 0.515$\pm$0.028 & 0.347$\pm$0.031 \\
BinMulti/L:1 & 0.728$\pm$0.112 & 0.692$\pm$0.194 \\
BinMulti/L:2 & 0.523$\pm$0.018 & 0.320$\pm$0.104 \\
BinCoral/B:1 & 0.770$\pm$0.010 & 0.770$\pm$0.010 \\
BinCoral/B:2 & 0.536$\pm$0.009 & 0.280$\pm$0.013 \\
BinCoral/L:1 & 0.733$\pm$0.115 & 0.696$\pm$0.196 \\
BinCoral/L:2 & \textbf{0.551}$\pm$0.018 & 0.264$\pm$0.077 \\
\bottomrule
\end{tabular}
\end{table}

We also trained the same set of models on a training corpus where both the binary and 
multi-class target was always selected as the highest label assigned by any annotator ("when in
doubt, treat it as misogynist"). The results for this experiment are shown in Table~\ref{table:choosemax}. On this data, the single-head binary model performed best 
both with respect to accuracy and F1.0-macro, while the single-head multiclass classification
model performed best on the multiclass task with respect to F1.0-macro.

In order to get an impression for how a binary model could be used to indicate forums where 
moderator intervention may be necessary because of a high number of misogynist postings, moderators 
selected 6 forums of which 3 have been found to have a high observed rate of misogynist postings and
3 without. The model BinCoral/L:1 from Table~\ref{table:mostfreq} assigned a rate of 0.23, 0.17, and 0.27 of postings in each of the first 3 forums to the misogyny class but only a rate of 0.01, 0.02 and  0.07 of the second 3 forums, thus giving a good indication of which of the forums moderators should 
prioritize. Looking at the actual postings labeled to be misogynist shows that in addition to 
many postings which were correctly identified, there was also a high number of false positives. 
However, many of these false positives are texts which are related to topics often raised in 
misogynist comments. In future work, the corpus will be expanded by providing manual annotations
for the false positives detected in those evaluation runs. 
The code for all experiments is based on the FARM library\footnote{https://github.com/deepset-ai/FARM} and is available online\footnote{URL withheld for anonymous review}.

Additional future work will analyse the performance of the different classification models on 
comments based on source, annotator disagreement or the kind of misogyny and improve the approach both
by further improving and extending the corpus and the models used for training. 

This work was conducted as part of project FemDwell\footnote{\url{https://ofai.github.io/femdwell/}} supported through FemPower IKT 2018\footnote{\url{https://austrianstartups.com/event/call-fempower-ikt-2018}}.

\bibliographystyle{acl_natbib}
\bibliography{references} 

\end{document}